\documentclass[runningheads]{llncs}
\usepackage{xcolor}
\usepackage{graphicx}
\usepackage{url}
\usepackage{multirow}
\usepackage{array}
\usepackage{booktabs}
\usepackage{enumitem}
\usepackage{float}
\usepackage{amssymb}
\usepackage{amsmath}
\usepackage{longtable}
\usepackage{pifont}
\usepackage{color}
\usepackage{cite}
\usepackage{subfigure}
\usepackage{changepage}
\usepackage[ruled,linesnumbered]{algorithm2e}
\usepackage[misc]{ifsym}

\begin{document}
\title{Efficient Policy Generation in Multi-Agent Systems via Hypergraph Neural Network}
\titlerunning{Actor Hypergraph Convolutional Critic Network}
\author{Bin Zhang \and Yunpeng Bai \and Zhiwei Xu \and Dapeng Li \and Guoliang Fan$^{(\textrm{\Letter})}$}
\authorrunning{Bin Zhang, Yunpeng Bai, et al.}
\tocauthor{Bin Zhang, Yunpeng Bai, Zhiwei Xu, Dapeng Li, Guoliang Fan}
\institute{Institute of Automation, Chinese Academy of Sciences. \\School of Artificial Intelligence, University of Chinese Academy of Sciences.\\
\email{\{zhangbin2020, baiyunpeng2020, xuzhiwei2019, lidapeng2020, guoliang.fan\}@ia.ac.cn}}
%
%
\maketitle              
\begin{abstract}
The application of deep reinforcement learning in multi-agent systems introduces extra challenges.
In a scenario with numerous agents, 
one of the most important concerns currently being addressed is how to develop sufficient collaboration between diverse agents. 
To address this problem, 
we consider the form of agent interaction based on neighborhood and propose a multi-agent reinforcement learning (MARL) algorithm based on the actor-critic method, which can adaptively construct the hypergraph structure representing the agent interaction and further implement effective information extraction and representation learning through hypergraph convolution networks, leading to effective cooperation.
Based on different hypergraph generation methods, we present two variants: Actor Hypergraph Convolutional Critic Network (HGAC) and Actor Attention Hypergraph Critic Network (ATT-HGAC).
Experiments with different settings demonstrate the advantages of our approach over other existing methods. 

\end{abstract}
\section{Introduction}
Intelligent decision-making problems has attracted a large number of academics in recent years because of its complexity and extensive application.
Deep reinforcement learning (DRL), 
which combines the function approximation capabilities of deep learning with the trial-and-error learning capabilities of reinforcement learning, 
is closer to real-world biological learning methods, 
and it has progressed quickly in many fields, 
yielding good study outcomes.
In the single-agent scenario, DRL's performance in Go \cite{Gogame} and Atari 2600 games \cite{Atari}, for example, has topped that of humans. Simultaneously, academics working on the intelligent decision-making issues in multi-agent systems have produced some impressive outcomes, including intelligent transportation system \cite{peng2021learning}, wireless sensor network management \cite{sharma2020distributed}, 
as well as Multiplayer Online Battle Arena (MOBA) and Real-Time Strategy (RTS) games \cite{StarCraft}.
However, there are still many obstacles in the multi-task and multi-agent setting that significantly restrict the algorithm's deployment and applicability in the real world.
When all agents are treated as a single entity, 
the joint action space grows exponentially with the number of agents \cite{marl}.
If each agent is individually trained through reinforcement learning, 
the Markov property of the environment will be invalid.
And because the environment is non-stationary, 
each agent has no way of knowing whether the reward it receives is the result of its own actions or those of others.

Therefore, finding creative training approaches and effectively extracting the attributes of agents is vital to lead the mutual cooperation of agents.
MADDPG \cite{MADDPG}, for example, employs a centralized training with decentralized execution structure to combine the benefits of the two methods.
Since the critic is only needed during the training phase, 
it is convenient to use all agents' input to develop a centralized critic for each independent actor, 
with each actor relying solely on its own local observations during the execution phase.
However, simply concatenate all agents' features may result in information redundancy.
Feature extraction and representation from high-dimensional and large-scale data can enhance agents' understanding of complex environments and improve their decision-making level, which is also crucial for MARL.
MAAC \cite{MAAC} leverages the attention mechanism \cite{attention} to get better results.
It allows agents to dynamically and selectively pay attention to the features of other agents.
The attention mechanism is also used by ATT-MADDPG \cite{att-maddpg} to complete the dynamic modeling of teammates. 
Furthermore, because agents in the system can naturally form graph topological structures depending on their locations, there has been a lot of work combining graph neural network (GNN) \cite{gnn} with MARL, such as DGN \cite{dgn} and MGAN \cite{mgan}.
However, the approaches described above need that each agent interact with all other agents in the system. In a complex environment, significant interaction between agents in a neighborhood is usually sufficient, whereas interaction between agents in different neighborhoods can be lessened.

To this end, we discuss the adaptive generation of neighborhoods in the multi-agent system and the cooperation of agents within and between neighborhoods.
We explore the application of the hypergraph neural network (HGNN) \cite{hgnn} in multi-agent reinforcement learning and propose Actor Hypergraph Convolutional Critic Network (HGAC) and Actor Hypergraph Attention Critic Network (ATT-HGAC). To achieve efficient state representation learning, the dynamic hypergraph is constructed adaptively and the hypergraph convolution is applied. Despite the complexity of the relationship between agents in the environment, our method is able to extract effective features from large amounts of information to achieve efficient strategy learning.
Experiments with different reward settings and different types of collaboration show that our approach outperforms other baselines. And the algorithm's working mechanism is revealed by ablation testing and visualization studies.

\section{Preliminaries}
\subsection{Markov Game}
We employ the framework of Markov Games (also known as Stochastic Games, SG) \cite{sg}, 
which is widely used as a standardized game model for sequential decision-making problems in multi-agent systems and can be seen as a multi-agent extension of the single-agent Markov Decision Process (MDP). 
It is represented by a tuple $\langle \mathcal{S},\mathcal{A}_1,..,\mathcal{A}_N,r_{1},...,r_{N},\mathcal{P,\gamma} \rangle$,
where $N$ is the number of agents and $\mathcal{S}$ is the environment state shared by all agents; 
$\mathcal{A}_i$ is the action set of agent $i$ and the joint action of all agents is described as $\mathcal{A}=\mathcal{A}_1 \times ...\times \mathcal{A}_N$.
If agent $i$ performs action $a$ in state $s$ and then transitions to new state $s'$, 
the environment will reward it with $r_i:\mathcal{S} \times \mathcal{A}_i \times \mathcal{S} \rightarrow \mathbb{R}$;
the new state $s'$ is determined by the state transition probability $\mathcal{P}:\mathcal{S \times A \times S} \rightarrow [0,1]$.
 Agent $i$ uses strategy $\mathbf{\pi}_i:\mathcal{S} \times \mathcal{A}_i\rightarrow[0,1]$ to take corresponding actions according to its current state,
 and the joint strategy of all agents is denoted as $\mathbf{\pi}=[\pi_1,...,\pi_N]$. 
 Following the conventional expression of game theory, 
 we use $(\pi_i,\pi_{-i})$ to distinguish the strategy of agent $i$ from all other agents. 
$\gamma$ represents the discount factor.
Under the framework of SG, all agents can move simultaneously in a multi-agent system.
If the initial state is $s$,
the value function of agent $i$ is expressed as the expectation of discounted return under the joint strategy $\mathbf{\pi}$: 
$v_{\pi_i,\pi_{-i}}(s)=\sum_{t\geq 0}\gamma^t \mathbb{E}_{\pi_i,\pi_{-i}} [r^j_t|s_0=s,\pi_i,\pi_{-i}] $.
According to the Bellman equation, 
the action-state value function can be written as: 
$Q_{\pi_{i}, \pi_{-i}}(s, \boldsymbol{a})=r_{i}(s, \boldsymbol{a})+\gamma \mathbb{E}_{s^{\prime} \sim p}\left[v_{\pi_{i}, \pi_{-i}}\left(s^{\prime}\right)\right]$.

\subsection{Hypergraph Learning} 
A hypergraph \cite{hypergraph} can be defined as $\mathcal{G}$ $ =(\mathcal{V, E})$, where $\mathcal{V}=\{v_1, ...,v_N\}$ denotes the set of vertices, $\mathcal{E} = \{\epsilon_1, ...,\epsilon_M\}$ denotes the set of hyperedges, 
$N$ and $M$ are the number of vertices and hyperedges, respectively. 
Unlike the edges in the graph, 
the hyperedge can connect any number of vertices in the hypergraph \cite{hypergraph}.
Hypergraph can be represented by an incidence matrix $\mathcal{H}\in \mathbb{R}^{N\times M}$, 
with elements specified as:
\begin{eqnarray}
h\left(v_{i}, \epsilon_{j}\right)=\left\{\begin{array}{ll}
1, & \text { if } v_{i} \in \epsilon_{j} \\
0, & \text { if } v_{i} \notin \epsilon_{j}
\end{array} \right.
\end{eqnarray}
where $v_i \in \mathcal{V}, \epsilon_{j} \in \mathcal{E}$.
Each hyperedge is given a weight $w_\epsilon$. 
All of the weights combine to produce a diagonal hyperedge weight matrix $\mathbf{W} \in \mathbb{R}^{M \times M}$.
In addition, the degrees of hyperedges and vertices are defined as $d(\epsilon)=\sum_{v\in V}h(v,\epsilon)$ and $d(v) = \sum_{\epsilon \in E}w_{\epsilon}h(v,\epsilon)$ respectively, which in turn constitute hyperedge diagonal degree matrix $\mathbf{D_e}$ and vertex diagonal degree matrix $\mathbf{D_v}$ respectively.

In a variety of domains, hypergraph learning is commonly employed. It was first used in semi-supervised learning methods as a propagation process \cite{zhou2006learning}. The learning of distinct modalities is handled in multi-modal learning by building different subhypergraphs and assigning weights \cite{zhu2015content}. In deep reinforcement learning, it is introduced to model the combined structure of multi-dimensional discrete action space and execute value estimation in a single-agent environment \cite{tavakoli2020learning}. In the value function decomposition method of multi-agent reinforcement learning, the utility function of each agent is fitted to the global action state value function using a hypergraph neural network \cite{hgmix}. Unfortunately, this method ignores the interaction between agents.

\section{Method}
In this section,  we introduce in detail our new methods called Actor Hypergraph Convolutional Critic Network (HGAC) and Actor Hypergraph Attention Critic Network (ATT-HGAC). 
We begin by discussing adaptive dynamic hypergraph generation. 
Secondly, we look at how hypergraphs are used in centralized critics to extract and represent information of agents. 
Finally, we give the overall MARL algorithm.

\begin{figure}[tbp]
    \centering
    \includegraphics[width=4.8 in]{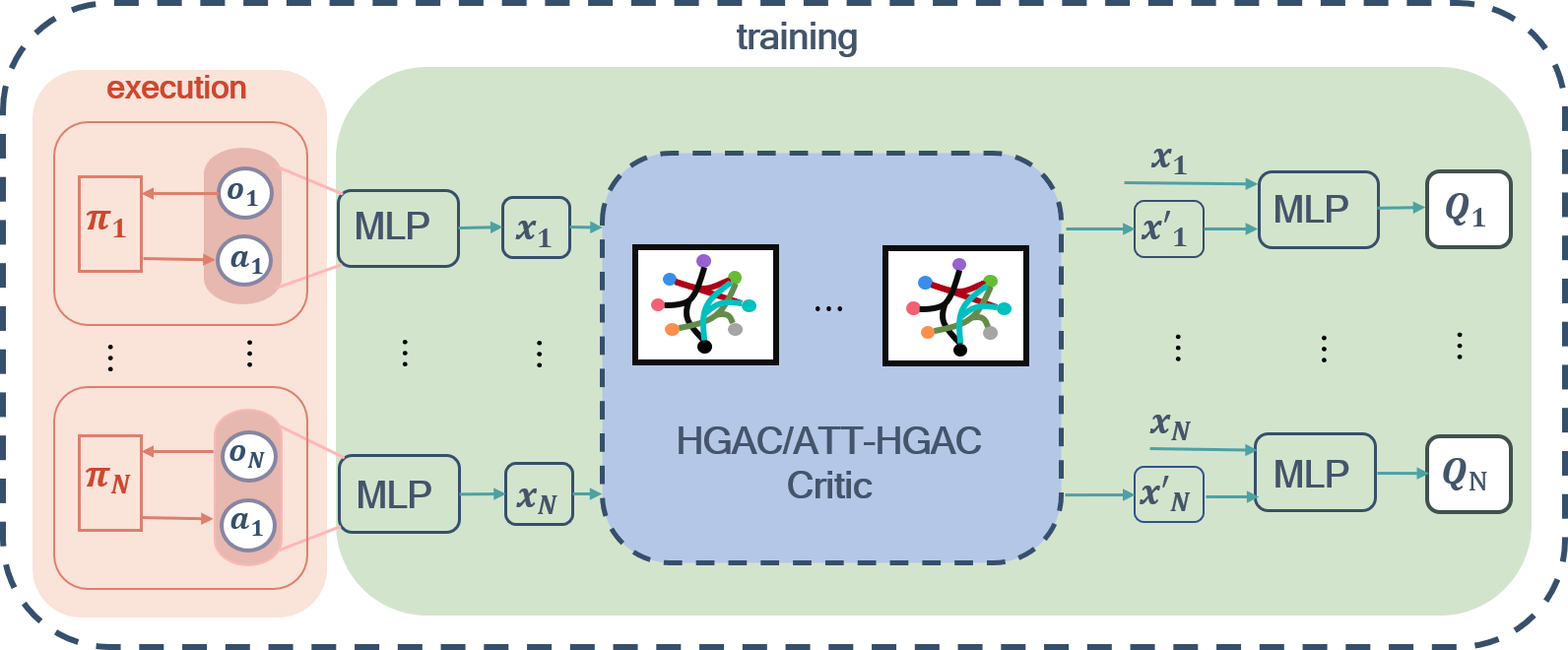}
    \caption{The overall architecture of HGAC/ATT-HGAC.}
    \label{fig:HGAC}
\end{figure}

\subsection{Hypergraph Generation}
Hypergraph, unlike the traditional graph structure, 
unites vertices with same attributes into a hyperedge.
In a multi-agent scenario, if the incidence matrix is filled with scalar $1$, 
as in other works' graph neural network settings, 
each edge is linked to all agents, 
then the hypergraph's capability of gathering information from diverse neighborhoods will be lost. 
Meanwhile, since the states of agents in a multi-agent scenario vary dynamically over time, 
the incidence matrix should be dynamically adjusted as well.

For the aforementioned reasons,
we investigate employing deep learning to dynamically construct hypergraphs.
And instead of using a 0-1 incidence matrix, 
we optimize the elements of the incidence matrix to values in the range of $[0,1]$,
which describe how strong the membership of the vertices in the hyperedge is.
In HGAC, we encode each agent's observation and action features and construct the agent's membership degree to each hyperedge using a Multilayer Perceptron (MLP) model:
\begin{eqnarray}
\mathbf{h}({v_i, E})=Softmax(MLP(concatenate(o_i,a_i))).
\end{eqnarray}
In addition, rather than utilizing the attention mechanism to aggregate neighbor information in MAAC, 
we propose using it to construct the hypergraph's incidence matrix.
However, calculating the attention weight of a hyperedge to a vertex is unusual since it presupposes that hyperedges are comparable to vertices.
To address this issue, we set the number of hyperedges equal to the number of vertices (agents) and assign each hyperedge to a specific agent. 
On each hyperedge, the membership degree of the specific agent is set to $1$.
Each hyperedge denotes a neighborhood of high-order attributes centered on the specific agent.
Using the attention mechanism to assess the similarity of other agents' attributes to its own, we can create the hypergraph's incidence matrix:
\begin{eqnarray}
h\left(v_{j}, \epsilon_{i}\right)=\left\{\begin{array}{ll}
1, & \text { if } i=j \\
\frac{\exp \left(f\left(x_{i}, x_{j}\right)\right)}{\sum_{m=1}^{N} \exp \left(f\left(x_{i}, x_{m}\right)\right)}, & \text { if } i \neq j
\end{array}\right.
\end{eqnarray}
where $x$ represents the feature of vertices, $f(x_i,x_j)$ is the score function used to calculate the correlation coefficient between \emph{query} and \emph{key}.
We define $f(x_i,x_j)= x_j^T W_k^T W_q x_i$, where $W_k$ and $W_q$ are learnable parameters as proposed in \cite{attention}.
Then we normalize the correlation coefficient by $softmax$ to obtain the attention coefficient of $i$ to $j$.

Based on the current observation and action features of all agents, 
the hypergraph generation network can adaptively generate various hyperedges.
Each hyperedge indicates a neighborhood with same or similar high-order features. 
It imply that agents on a hyperedge are in close proximity, or that agents have the same action intention, and so on.

\begin{figure}[tbp]
    \centering
    \includegraphics[width=4.8 in]{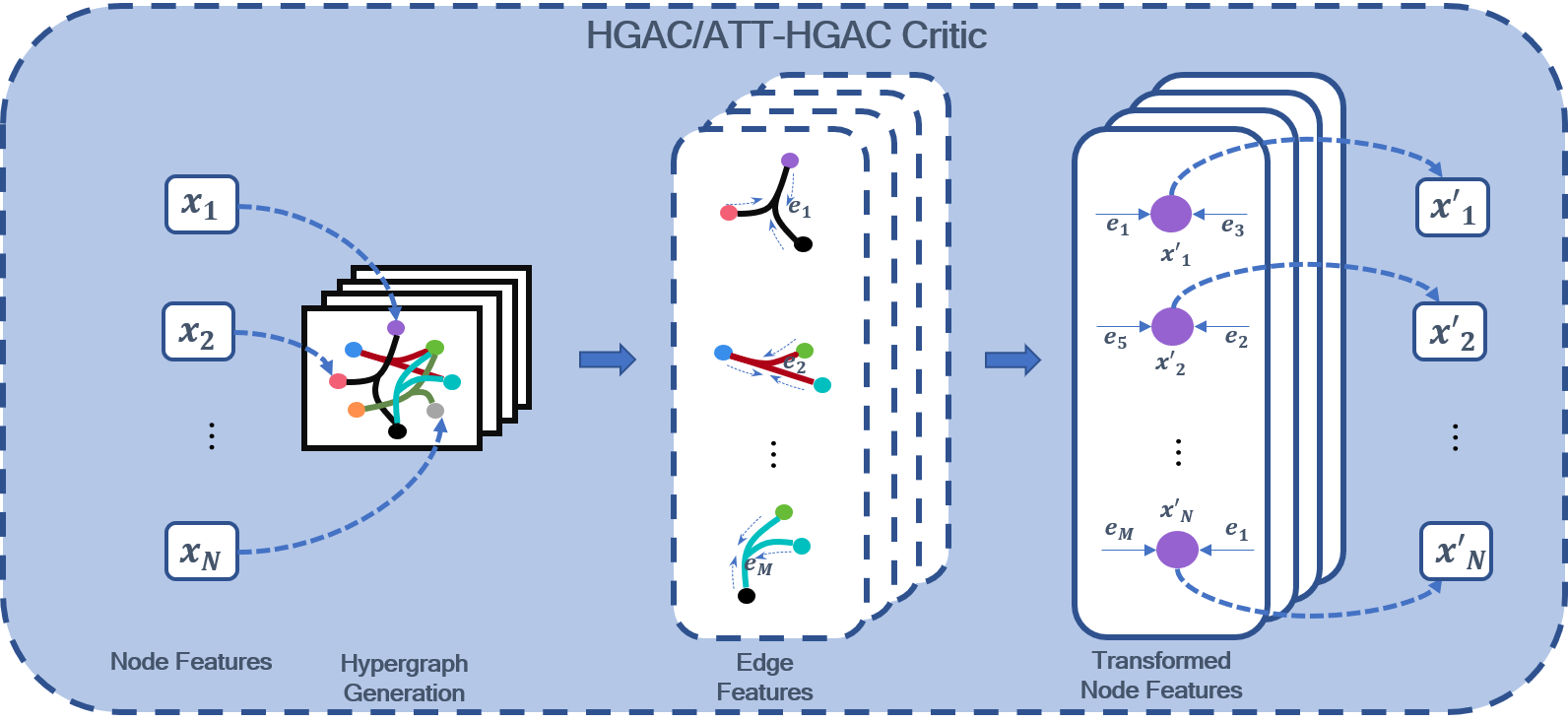}
    \caption{The overall architecture of HGAC/ATT-HGAC. Each agent's feature $h_i$, including observation and action, is used to construct a dynamic hypergraph structure, with each hyperedge $e_i$ representing a neighborhood. In the hypergraph convolution process, the agents in the same neighborhood aggregate information to the hyperedge feature to realize neighborhood cooperation and interaction. Subsequently, the embedding of each agent $h'_i$ is aggregated by the hyperedge information to realize cooperative interaction between different neighbors.}
    \label{fig:HGC}
\end{figure}

\subsection{Hypergraph Convolution Critics}

Following the generation of the hypergraph, 
hypergraph neural networds can be used to train the centralized critics to guide the optimization of decentralized execution strategies, 
allowing the agents on same hyperedges to achieve strong coordination and agents on different hyperedges to realize weak coordination.
To train agents' new feature embedding vectors,
we employ a two-layer hypergraph convolutional network. 
Referring to the HGNN's convolution formula \cite{hgnn}, the hypergraph convolution operator is defined as:
\begin{eqnarray}
\mathbf{x^{(l+1)} =  \sigma (D_{v}^{-1 / 2} H W D_{e}^{-1} H^{\top} D_{v}^{-1 / 2} x^{(l)} P^{(l)}}),
\end{eqnarray}
where $\mathbf{W}$ and $\mathbf{P}$ as learnable parameters represent the hyperedge weight matrix and the linear mapping of the vertices features, respectively.
In a convolution process, 
vertices with the same high-order feature attributes combine their information into the hyperedges to which they belong to generate hyperedges feature vectors. 
After that, each agent's feature representation will be weighted and aggregated from the hyperedge's feature to which it belongs.
Furthermore, as indicated in Figure~\ref{fig:HGC}, 
we create several hypergraph convolutional neural networks simultaneously to aid the algorithm in gaining a better understanding of crucial information and increasing its robustness.
The new characteristics received by each vertex after the convolution operation fuse all of the vertices features required for the agent to collaborate,
but naturally, its original attributes are smoothed out.
Inspired by DGN \cite{dgn}, 
we connect the features of original vertices and new features generated by each head of hypergraph convolutional networks and input them into the critic network.
The Q-value function of agent $i$ is calculated by:
\begin{eqnarray}
Q_i=ReLU(MLP(concatenate(x_i,x_{i_1}',...,x_{i_K}')),
\end{eqnarray}
where $x_i$ is the initial feature embedding of agent $i$,
and $(x_{a_1}'$$,...,x_{a_K}')$ is the new feature embedding generated by hypergraph convolution of $K$ heads.
In addition, all parameters of feature embedding and critic networks are shared, 
considerably reducing training complexity and increasing training efficiency.

\subsection{Learning with Hypergraph Convolution Critics}
To stimulate agent exploration and prevent converging to non-optimal deterministic policies, we advocate employing maximum entropy reinforcement learning \cite{sac} for training.
In addition, unlike the settings in MADDPG \cite{MADDPG}, parameter sharing allows us to update all critics together. 
The loss function of critic networks is defined as:
\begin{eqnarray}
\begin{array}{c}
\mathcal{L}(\theta)=\sum\limits_{i=1}^{N} \mathbb{E}_{\left(o, a, r, o^{\prime}\right) \sim D}\left[\left(Q_{i}^{\theta}(o, a)-\mathop{target}_{i}\right)^{2}\right],
\end{array}
\end{eqnarray}
where $o$ represents the observation of the agent, $D$ is the reply buffer used for experience reply, other symbol settings are the same as in Markov Games, and $target_{i}=r_{i}(o,a)+\gamma \mathbb{E}_{a^{\prime} \sim \pi_{\bar{\mu}}\left(o^{\prime}\right)}\left[Q_{i}^{\bar{\theta}}\left(o^{\prime}, a^{\prime}\right)-\right. 
\left.\omega \log \left(\pi_{\bar{\mu}_{i}}\left(a_i^{ \prime} \mid o_i^{\prime}\right)\right)\right]$.
$Q_{i}^{\theta}(o, a)$ is the Q-value function(parameterized by $\theta$), $target_i$ is the target Q-value function which is calculated by environmental rewards $r_i$, target critics $Q_{i}^{\bar{\theta}}$ (parameterized by $\bar \theta$) and target policies $\pi_{\bar{\mu}_{i}}$(parameterized by $\bar \mu$). 
$\omega$ is a temperature coefficient to balance the maximization of entropy and rewards.
In terms of actor networks, the policy gradient of each agent is expressed as:
\begin{eqnarray}
\begin{aligned}
\nabla_{\mu_{i}} J\left(\pi_{\mu}\right)=& \mathbb{E}_{o \sim D, a \sim \pi}\left[\nabla_{\theta_{i}} \log \left(\pi_{\mu_{i}}\left(a_{i} \mid o_{i}\right)\right) \cdot\right.\\
&\left.\left(-\omega \log \left(\pi_{\mu_{i}}\left(a_{i} \mid o_{i}\right)\right)+A\left(o, a_{-i}\right)\right)\right].
\end{aligned}
\end{eqnarray}
Inspired by COMA \cite{coma}, we use the advantage function $A(o,a_{-i})=Q_{i}^{\theta}(o, a)-\mathbb{E}_{a_{i} \sim \pi_{i}\left(o_{i}\right)}\left[Q_{i}^{\theta}\left(o,\left(a_{i}, a_{- i}\right)\right)\right]$ with a counterfactual baseline, which can achieve the purpose of credit assignment by fixing the actions of other agents and comparing the value function of a specific action with the expected value function so as to determine whether the action lead to an increase or decrease in the expected return.

The whole algorithm adopts the framework of centralized training with decentralized execution (CTDE) \cite{foerster2016learning} and its structure is shown in Figure\ref{fig:HGAC}.
It extract the features of agents within and between neighborhoods to guide their value function estimation during training process,
so that agents can only follow their own observations during the actual execution process and do not require any other input to complete complex collaboration strategies.

\section{Experiments}

\begin{figure*}[tbp]
\centering
\subfigure[Cooperative Treasure Collection(CTC)]{
    \includegraphics[width = 1.4 in]{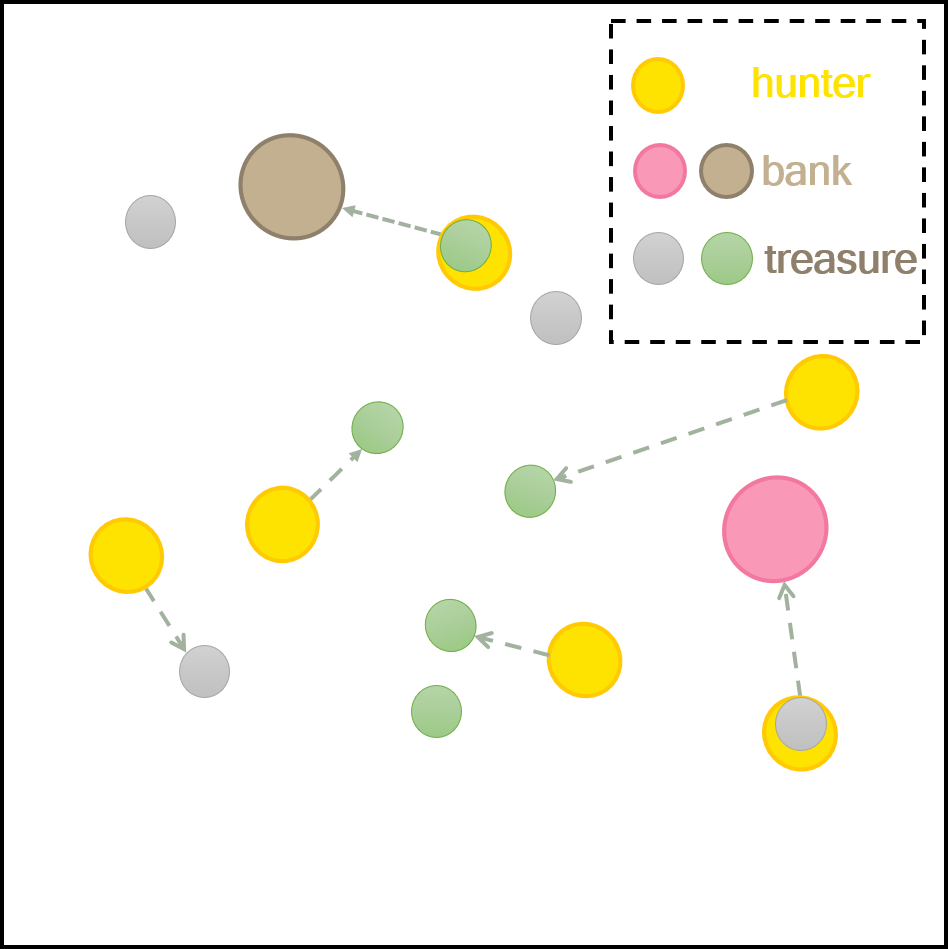}
}
\hspace{0.2cm}
\subfigure[Rover-Tower(RT)]{
    \includegraphics[width = 1.4 in]{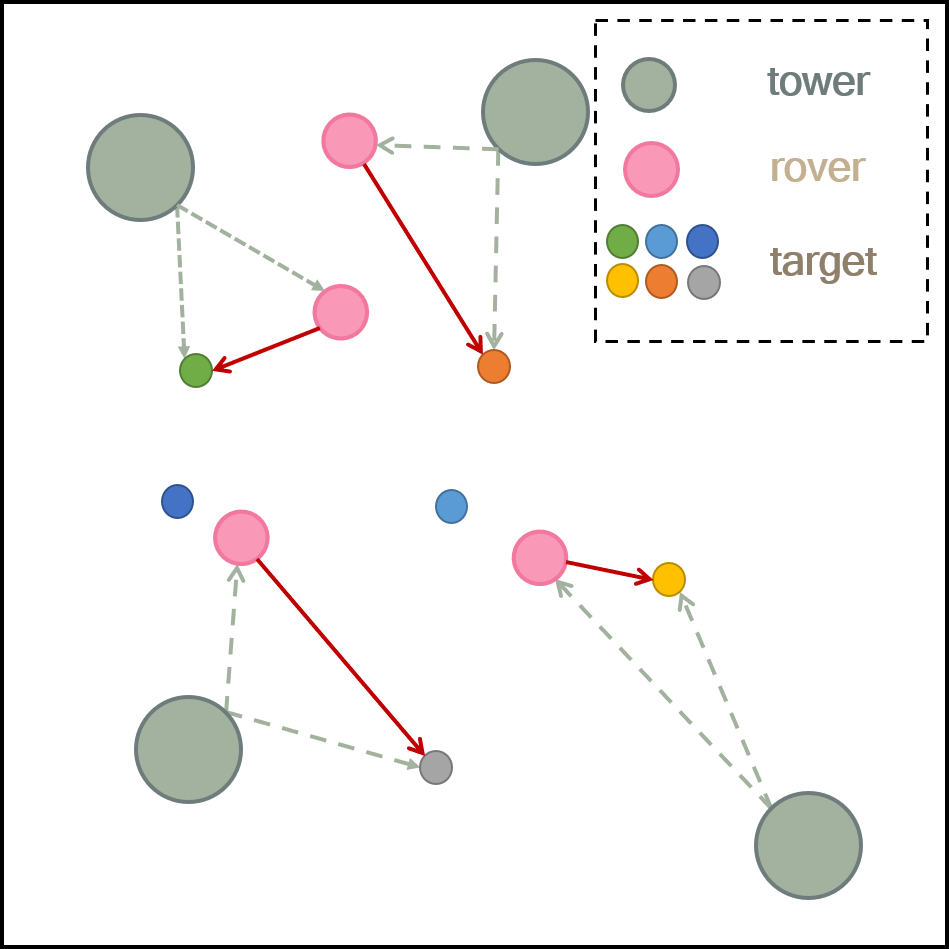}
}
\hspace{0.2 cm}
\subfigure[Cooperative Navigation(CN)]{
    \includegraphics[width = 1.4 in]{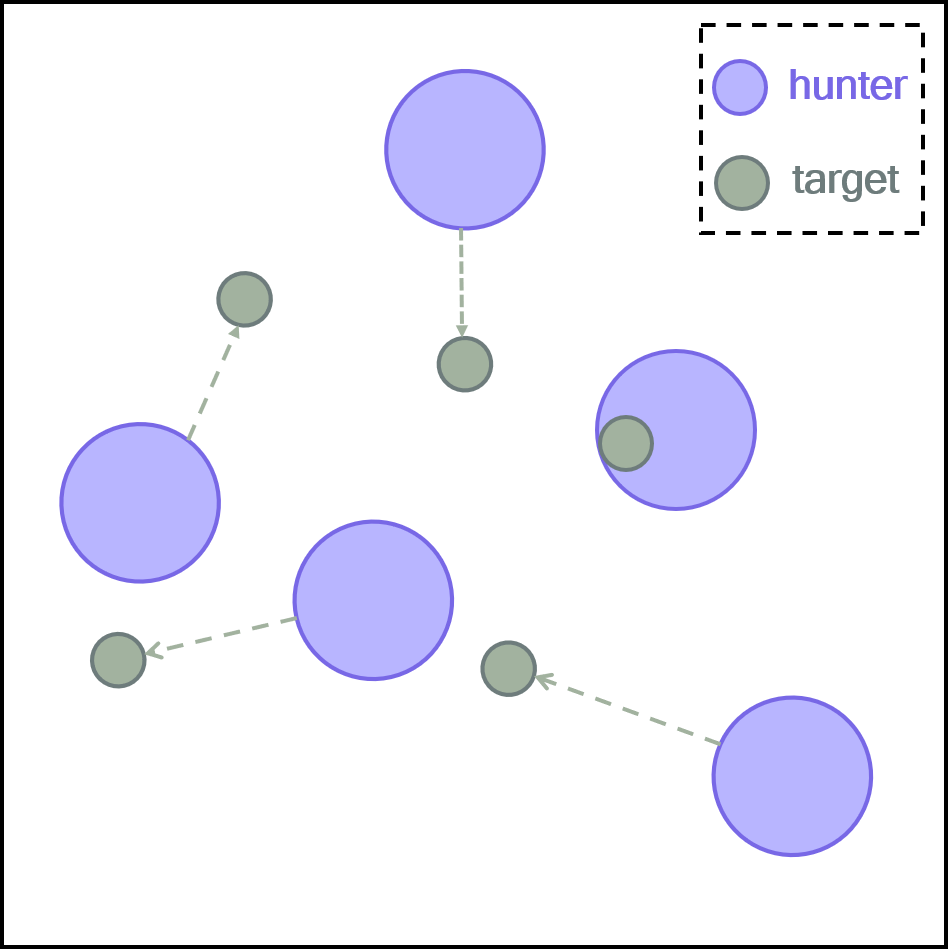}
}
\hspace{0.2cm}
\caption{multi-agent particle environments used for our evaluating. In CTC and CN, dotted lines point to the target to which the agent needs to go. In RT, dotted lines refers to the rover and target declared by tower, and solid lines indicates the rover's target.}
\label{scenario}
\end{figure*}

Multi-agent particle environment (MPE) \cite{MADDPG} is one of the most commonly used tasks to evaluate MARL algorithms. 
It simplifies environment animation while still allowing for some basic physical simulation, and it focuses on evaluating strategy effectiveness.
In this section, we evaluate HGAC/ATT-HGAC and other baselines in scenarios of multi-agent particle environments with different observation and reward settings, and investigate the algorithm mechanism through ablation and visualization researches.

\subsection{Settings}
We consider environments with continuous observation spaces and discrete action spaces. Specifically, considering different agent types and reward settings, we use three benchmark test environments, including Cooperative Treasure Collection (CTC) and Rover-Tower (RT) proposed in MAAC, and Cooperative Navigation (CN) introduces in MADDPG. They are shown in Figure \ref{scenario}.

\noindent
\textbf{Cooperative Treasure Collection. }
6 hunters are in charge of gathering treasures, while 2 banks are in charge of keeping treasures, with each bank only storing treasures of a specific hue.
Hunters will be rewarded for acquiring treasures on an individual basis. 
No matter who successfully deposits treasures in the proper bank, all agents will earn global rewards, and if they collide, they will be penalised.

\noindent
\textbf{Rover-Tower. }
4 rovers, 4 towers and 4 landmarks. Rovers and towers are paired randomly in each episode. The tower sends the location signal of the landmark to its paired rover, then the rover receives the signal and heads to the destination. The rewards of each pair are determined by the distance between the rover and the destination.

\noindent
\textbf{Cooperative Navigation. }
5 hunters, 5 landmarks. Hunters need to work together to cover all landmarks and avoid collisions. 
The environmental rewards are determined by the distance between hunters and landmarks and whether there are collisions.

We choose the well-known MARL methods MADDPG and MAAC as well as completely decentralized independent learning methods DDPG \cite{ddpg} and SAC \cite{sac} as baselines to compare with our proposed HGAC/ATT-HGAC approach.
Since DDPG and MADDPG are algorithms proposed under continuous control scenarios, 
we apply the gumbel-softmax reparameterization trick \cite{MADDPG} to deal with discrete action scenarios.
Furthermore, in order to focus on the enhancement of the experimental effect on the hypergraph convolutional critic network and reduce the impact of the underlying reinforcement learning method, we implement an additional SAC algorithm based on the CTDE framework and named it MASAC.
The hyperparameters common to all algorithms remain the same.
We examine the performance of HGAC in the Cooperative Treasure Collection and Cooperative Navigation scenarios, as well as the performance of ATT-HGAC in Rover-Tower, 
due to characteristics of the experimental environments.

All environments are trained with 60000 episodes,
with each episode having 25 time steps and the program having 12 parallel rollouts.
During the training process, we keep track of the average return of each episode.

\subsection{Results and Analysis}
\begin{figure*}[htbp]
\centering
\subfigure[CTC]{
    \includegraphics[width = 1.52 in]{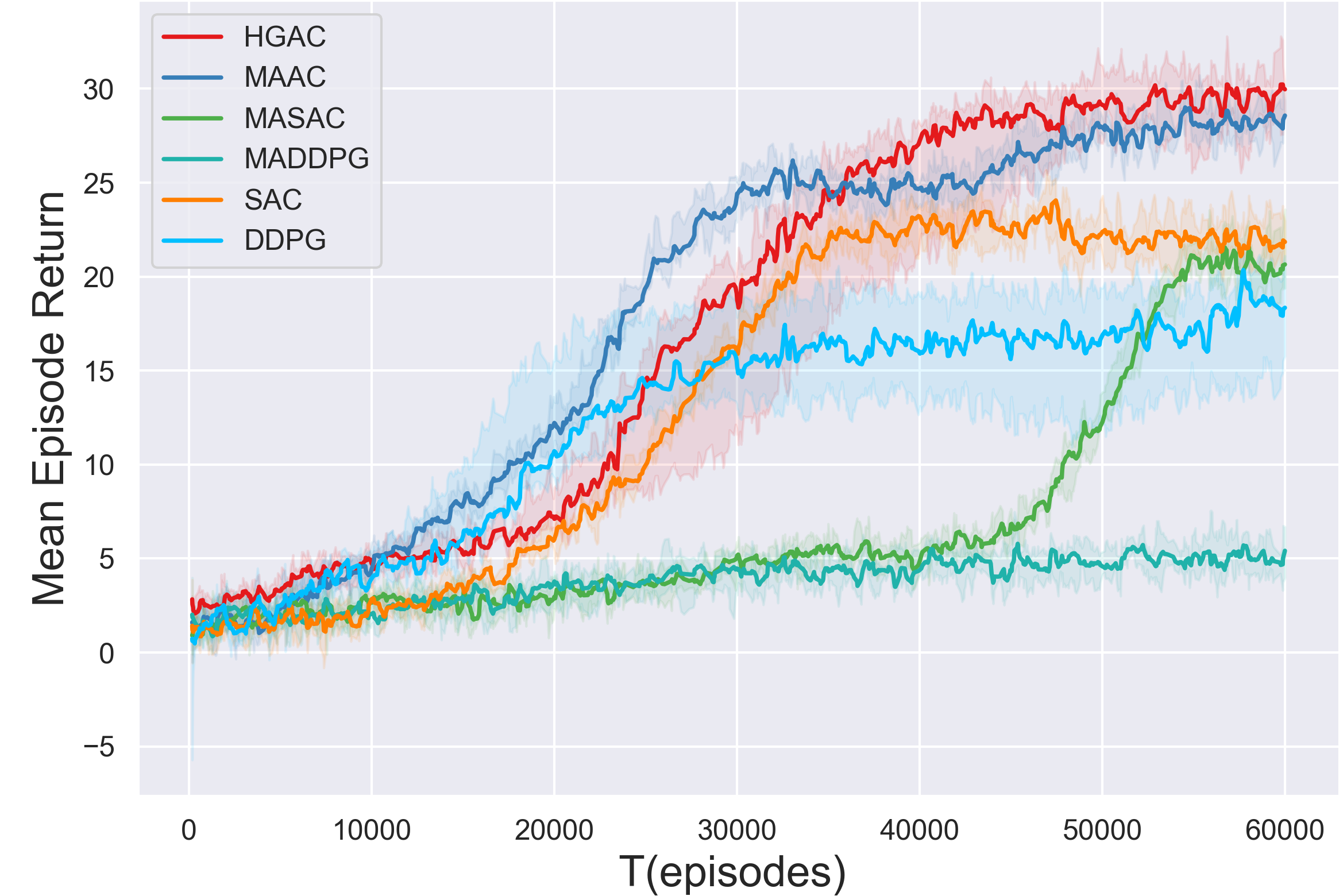}
}
\hspace{-0.2cm}
\subfigure[RT]{
    \includegraphics[width = 1.5 in]{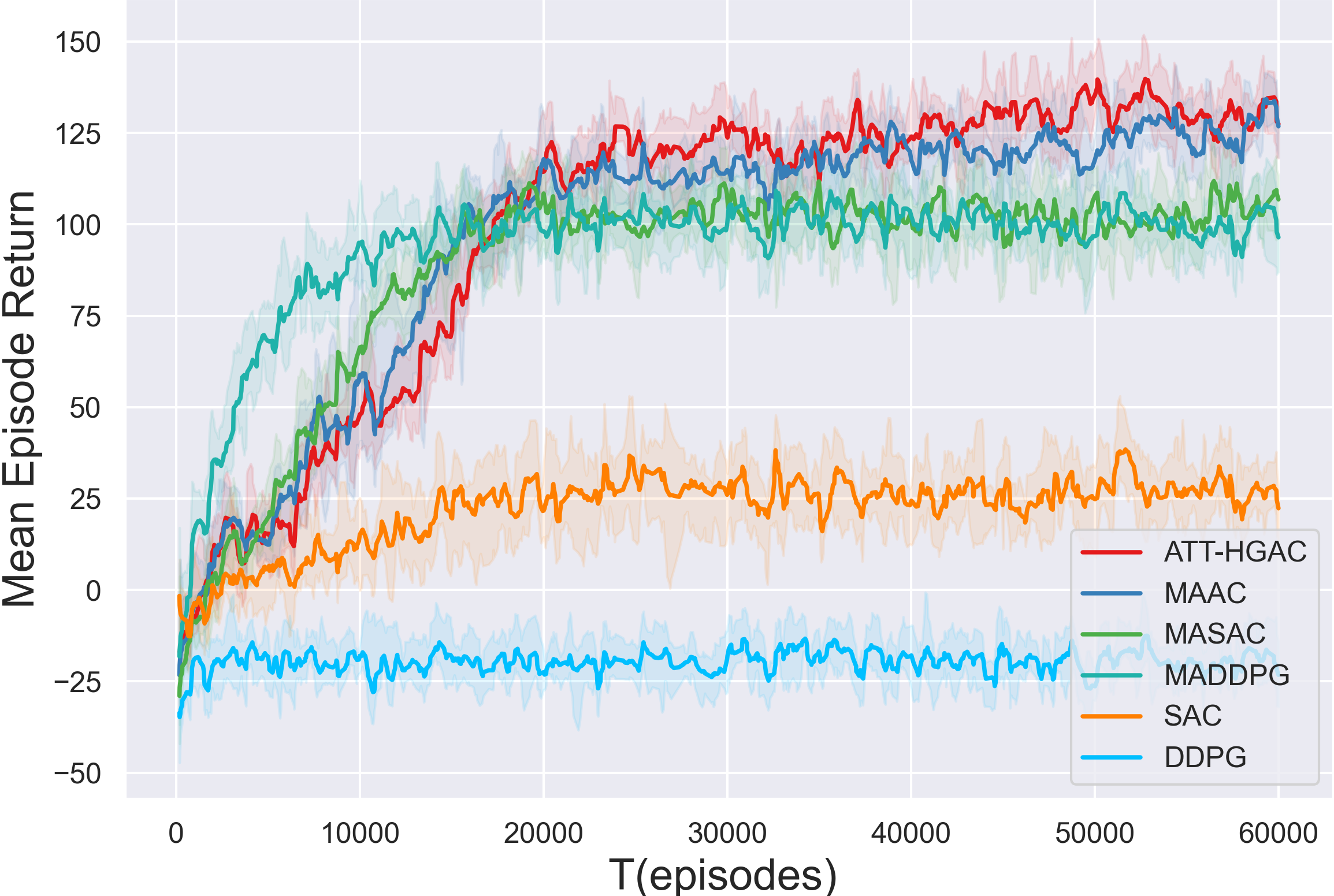}
}
\hspace{-0.2 cm}
\subfigure[CN]{
    \includegraphics[width = 1.5 in]{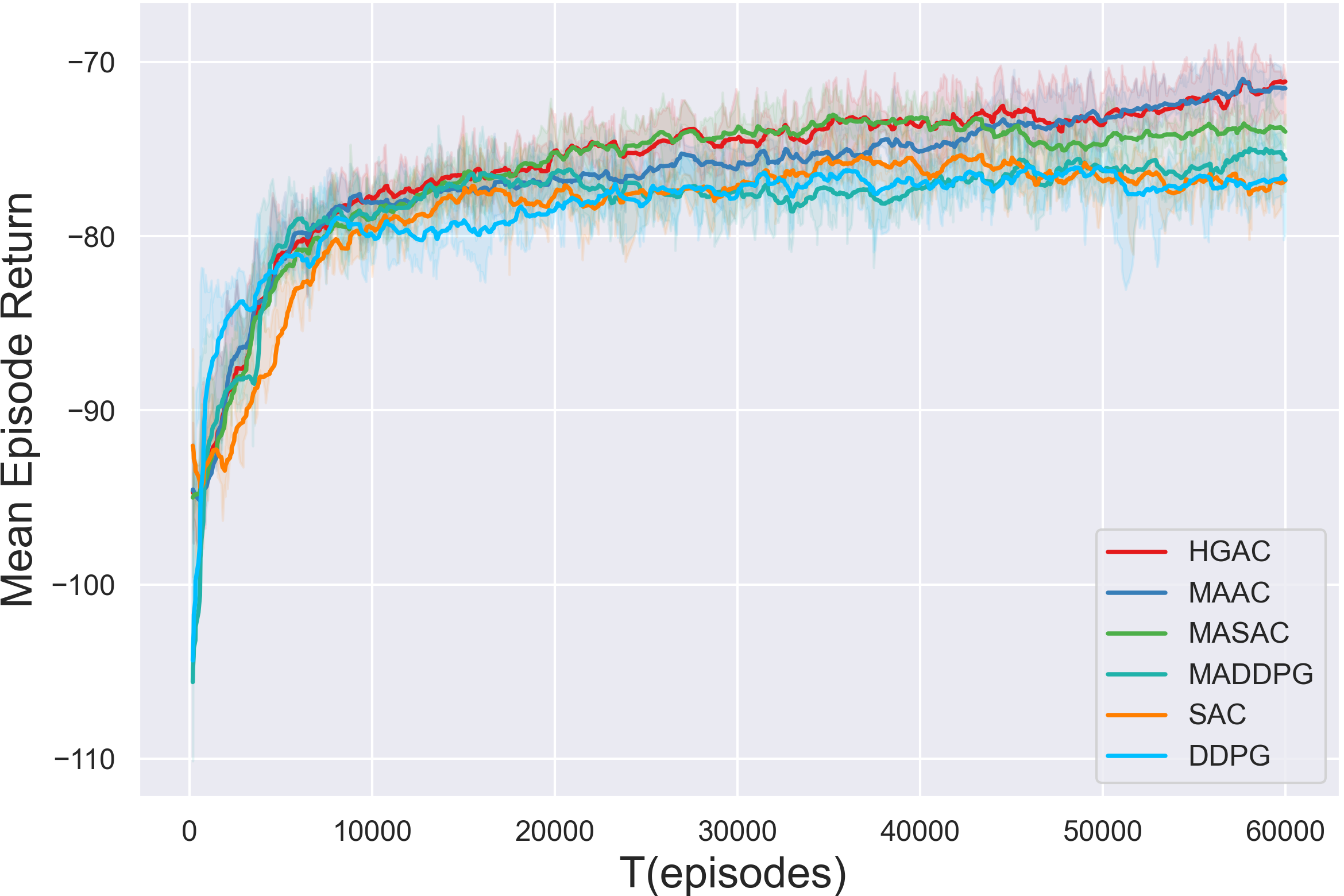}
}
\hspace{-0.2cm}
\caption{Performance curves with HGAC/ATT-HGAC, MAAC, MADDPG, MASAC, DDPG, SAC for 3 multi-agent particle environments.The solid line represents the median return, and the shadow part denotes the standard deviation. }
\label{results}
\end{figure*}
Figure \ref{results} illustrates the average return of each episode obtained by all algorithms with five random seeds tests in three environments.
The results reveal that our methods are quite competitive when compared to other algorithms.

\subsubsection{Experimental Results }
The results show that using SAC as the underlying algorithm has smaller variance and better performance than DDPG.
In the CN scenario, since the task is relatively simple, all algorithms achieve good results.
But on the whole, MARL methods have better performance than the fully decentralized methods.
And it is not hard to see that our HGAC have advantages over other algorithms.

In the CTC scenario, 
although different types of agents have different reward and observation settings,
the convergence result of HGAC is surprising.
Single-agent RL algorithms can even yield decent results in this circumstance since all agents can acquire global state information.
In contrast, MADDPG and MASAC, which merely concatenating all of the agents' information as the input of the critic networks, perform badly due to the input of excessively redundant data and the lack of feature extraction capability.
Correspondingly, 
HGAC can adaptively split high-order attribute neighborhoods, achieve strong cooperation inside the neighborhood and weak collaboration between neighborhoods to obtain the best performance.

In the RT scenario, ATT-HGAC also performs at an excellent level.
Single-agent reinforcement learning algorithms are completely ineffective since rovers' local observation is $0$ and they can only receive discrete signals of all targets given by all towers.
ATT-HGAC that uses the attention mechanism to create hypergraph enable rovers to focus on information from their own signal towers and achieve better outcomes than MADDPG/MASAC.

\subsubsection{Ablation studies}
\begin{figure}[htp]
    \centering
    \includegraphics[width=0.7\textwidth]{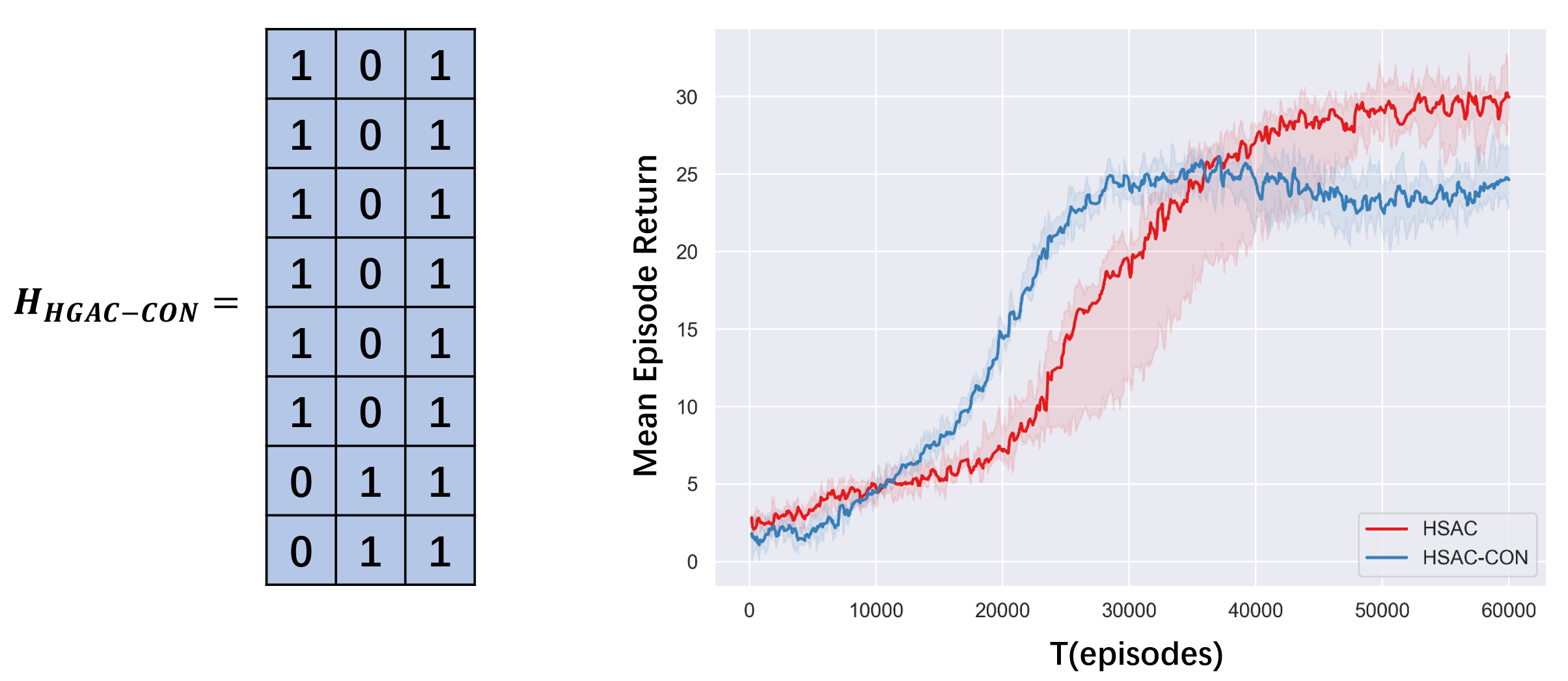}
    \caption{\textit{Left}:Incidence matrix constructed using prior knowledge. \textit{Right}:Performance curve of HGAC and HGAC-CON.}
    \label{ablation}
\end{figure}


To assess the effectiveness of the hypergraph generating technique, we performe an ablation experiment. 
We create a static hypergraph based on prior knowledge and utilize it to train the critic network.
In the CTC environment, specifically, 
six hunters are connected using one hyperedge, two banks are connected using another hyperedge, and all agents are connected together using a third hyperedge.
We keep other settings the same as HGAC and name it HGAC-CON.

Figure \ref{ablation} shows the final experimental result. 
Although HGAC-CON has a faster convergence rate than HGAC,
its ultimate performance is inferior to that of the HGAC.
This is pretty simple to comprehend.
The use of prior knowledge allows the algorithm to skip the stage of hypergraph generation,
which speeds up the effect but restricts the hypergraph's expressiveness.
As a result, self-adaptive dynamic hypergraph generation has the potential to generate better results.

\subsubsection{Visualization Research}
\begin{figure}[htp]
    \centering
    \includegraphics[width=0.78\textwidth]{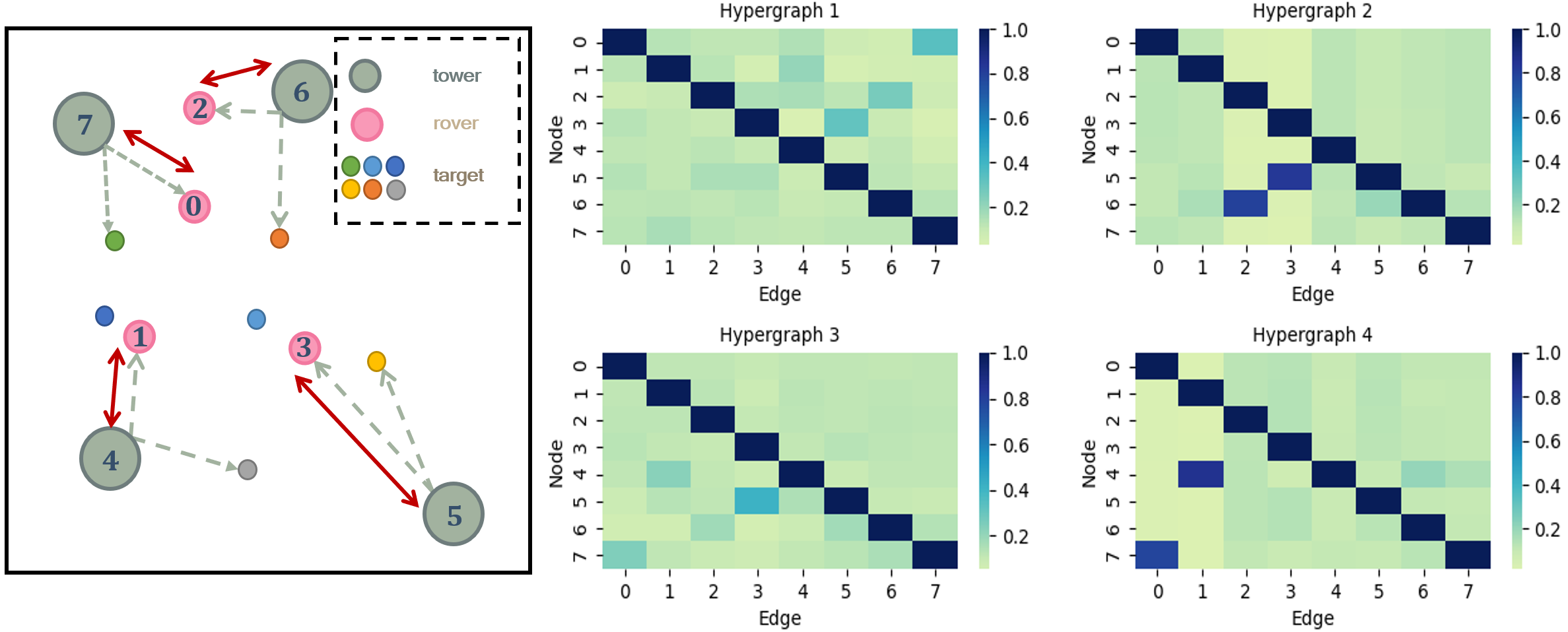}
    \caption{\textit{Left}: Correspondence between towers and rovers. \textit{Right}: Incidence matrixes heat map generated by ATT-HGAC.}
    \label{visualization}
\end{figure}
We perform a visualization experiment on the hypergraphs generation to investigate the effect of applying the attention mechanism to generate hypergraphs in ATT-HGAC.
As we hope, in the absence of clear supervision signals, rovers on different hyperedges successfully find signal towers they need to listen to.
As shown in Figure \ref{visualization}, agents 0-3 indicate rovers and agents 4-7 represent towers.
In the \emph{Hypergraph 1}, all of the towers (edge 4-7) successfully notice their rovers (node 0-3). 
And rovers also successfully notice their corresponding towers in the remaining three hypergraphs.
Four hypergraphs can learn the same pairings, they proves and complements each other.

\section{Conclusion and Future Work}
In this paper, we propose HGAC/ATT-HGAC, a novel method for applying hypergraph convolution to the centralized training with decentralized execution paradigm.
Our key contribution is to model agents adaptively as hypergraph structures, implement adaptive partition of neighborhoods,
as well as efficient information feature extraction and representation to aid actors in forming more effective cooperative policies. 
We evaluate our algorithms' performance in several multi-agent test scenarios, 
including various observation and rewards settings. 
The ablation experiment and visualization verify our method's efficacy and the importance of each component.
Facts have proved that HGAC/ATT-HGAC can successfully extract high-order neighborhood information to lead agents to attain efficient collaboration.
In the future, we consider making full use of the structural advantages of hypergraphs to carry out related research in the field of multi-agent communication, 
while improving the efficiency of the algorithm, and increasing its convergence speed and scalability.

%
%
%
%

\end{document}